\documentclass{article}

\usepackage{arxiv}
 \usepackage{amsmath} 
\usepackage[utf8]{inputenc} 
\usepackage[T1]{fontenc}    
\usepackage{hyperref}       
\usepackage{url}            
\usepackage{booktabs}       
\usepackage{amsfonts}       
\usepackage{nicefrac}       
\usepackage{microtype}      
\usepackage{lipsum}		
\usepackage{graphicx}
\usepackage{doi}
\usepackage{multicol}
\usepackage{multirow}
\usepackage{tikz,graphics,color,fullpage,float,epsf,caption,subcaption}

\title{Out-of-Distribution Detection and Data Drift Monitoring using Statistical Process Control}

\author{ \href{https://orcid.org/0000-0003-4723-5539}{\includegraphics[scale=0.06]{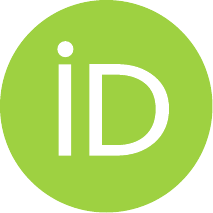}\hspace{1mm}Ghada Zamzmi}
\\
	Office of Science and Engineering Laboratories\\
	Center for Devices and Radiological Health\\
    U.S. Food and Drug Administration\\
	Silver Spring, MD 20993\\
	\And
	\href{https://orcid.org/0000-0001-8153-0115}{\includegraphics[scale=0.06]{orcid.pdf}\hspace{1mm}Kesavan Venkatesh} \\
	Department of Biomedical Engineering \\ Johns Hopkins University\\
	Baltimore, MD 21205 \\
 \And
	\href{https://orcid.org/0000-0001-9213-3131}{\includegraphics[scale=0.06]{orcid.pdf}\hspace{1mm}Brandon Nelson} \\
	Office of Science and Engineering Laboratories\\
	Center for Devices and Radiological Health\\
    U.S. Food and Drug Administration\\
	Silver Spring, MD 20993\\
 \And
	\href{https://orcid.org/0000-0003-1616-1290}{\includegraphics[scale=0.06]{orcid.pdf}\hspace{1mm}Smriti Prathapan} \\
	Office of Science and Engineering Laboratories\\
	Center for Devices and Radiological Health\\
    U.S. Food and Drug Administration\\
	Silver Spring, MD 20993\\
  \And
	\href{https://orcid.org/0000-0003-2804-2264}{\includegraphics[scale=0.06]{orcid.pdf}\hspace{1mm} Berkman Sahiner} \\
	Office of Science and Engineering Laboratories\\
	Center for Devices and Radiological Health\\
    U.S. Food and Drug Administration\\
	Silver Spring, MD 20993\\
  \And
	\href{https://orcid.org/0000-0001-9433-8093}{\includegraphics[scale=0.06]{orcid.pdf}\hspace{1mm}Paul Yi} \\
	Medical Intelligent Imaging Center\\
	University of Maryland School of Medicine\\
 University of Maryland \\
	Baltimore, MD 21201\\
  \And
	\href{https://orcid.org/0000-0001-9957-2866}{\includegraphics[scale=0.06]{orcid.pdf}\hspace{1mm}Jana G. Delfino} \\
	Office of Science and Engineering Laboratories\\
	Center for Devices and Radiological Health\\
    U.S. Food and Drug Administration\\
	Silver Spring, MD 20993\\
}

\renewcommand{\shorttitle}{\textit{arXiv} Template}

\hypersetup{
pdftitle={A template for the arxiv style},
pdfsubject={q-bio.NC, q-bio.QM},
pdfauthor={David S.~Hippocampus, Elias D.~Striatum},
pdfkeywords={First keyword, Second keyword, More},
}

\begin{document}
\maketitle

\begin{abstract}
\textbf{Background}: Machine learning (ML) methods often fail with data that deviates from their training distribution. This is a significant concern for ML-enabled devices in clinical settings, where data drift may cause unexpected performance that jeopardizes patient safety. 

\textbf{Method}: We propose a ML-enabled Statistical Process Control (SPC) framework for out-of-distribution (OOD) detection and drift monitoring. SPC is advantageous as it visually and statistically highlights deviations from the expected distribution. To demonstrate the utility of the proposed framework for monitoring data drift in radiological images, we investigated different design choices, including methods for extracting feature representations, drift quantification, and SPC parameter selection.

\textbf{Results}: We demonstrate the effectiveness of our framework for two tasks: 1) differentiating axial vs. non-axial computed tomography (CT) images and 2) separating chest x-ray (CXR) from other modalities. For both tasks, we achieved high accuracy in detecting OOD inputs, with $0.913$ in CT and $0.995$ in CXR, and sensitivity of $0.980$ in CT and $0.984$ in CXR. Our framework was also adept at monitoring data streams and identifying the time a drift occurred. In a simulation with 100 daily CXR cases, we detected a drift in OOD input percentage from $0-1\%$ to $3-5\%$ within two days, maintaining a low false-positive rate. Through additional experimental results, we demonstrate the framework's data-agnostic nature and independence from the underlying model's structure.

\textbf{Conclusion}: We propose a framework for OOD detection and drift monitoring that is agnostic to data, modality, and model. The framework is customizable and can be adapted for specific applications.

\end{abstract}

\keywords{Out of distribution detection \and drift monitoring \and statistical processing control \and medical imaging}

\section{Introduction}
Artificial intelligence (AI) is becoming prevalent in all areas of our lives, including healthcare. AI models have shown a strong ability to identify complex patterns in large datasets. However, the strength of a model's ability to learn from source data is also its weakness, as that data-dependence means performance can vary drastically--or the model can fail catastrophically-- when exposed to data that exhibits characteristics or patterns not represented in the training dataset. 

In AI systems applied to healthcare, the performance of the machine learning (ML) model can change over time or across different conditions due to shifts in patient demographics, updates in data acquisition technology, receipt of incorrect data types, or wrong labels \cite{sahiner_BJR_2023}. Such changes in the input data can substantially influence the model's safety and effectiveness and can lead to unreliable or erroneous predictions. This challenge is common in supervised machine learning models, which are often designed to classify inputs into predetermined classes and may struggle with unknown or irrelevant inputs \cite{ge2021evaluation,koch2023lord}. For instance, a model trained to classify chest x-rays (CXR) as normal or pneumonia might inaccurately categorize an unrelated knee x-ray due to its design constraints. In fact, this limitation of supervised ML classifiers led to the development of an entire area of machine learning research known as open-world classification \cite{geng2020recent}, which focuses on extending static or closed models to reject unknown images.

Fundamentally, `monitoring' refers to the ongoing observation of a process and its quality over time. Monitoring can focus on model inputs or outputs. Monitoring outputs is needed to ascertain consistency, quality, and reliability of the model's predictions. Scrutinizing outputs can reveal potential biases, inaccuracies, or evolving challenges that might arise post deployment. 
Monitoring inputs is equally important, as it informs whether the model is receiving the right data for processing (quality control) and can flag for input data drift due to changes in patient population or updates in data acquisition technology. Without input monitoring, there is a risk of the model processing out-of-distribution (OOD) inputs, which can lead to unpredictable outputs. For instance, a study \cite{magudia2021trials} reported the high frequency of mislabeling errors within CT exams, which could potentially lead to the misapplication of ML models or training of ML models with inappropriate data, highlighting a tangible challenge in the deployment of medical imaging AI. For these reasons, Feng et al. \cite{Feng_Nature_2022} discussed the need to establish approaches for the active monitoring of models deployed in clinical environments.

In this paper, we propose a framework for OOD detection and input drift monitoring that combines ML methods and geometric distances with SPC methods. This work includes a detailed exploration of essential factors such as feature selection, metrics, and various monitoring options. The ability to recognize OOD and monitor input not only enhances a model's performance, but also minimizes trust-eroding failures, underlining the critical role of thorough and effective monitoring in ML applications \cite{ge2021evaluation,koch2023lord}.

\subsection{Statistical Process Control}
Statistical process control (SPC) is a statistical method to monitor and control the quality of a process. SPC improves process quality through a better understanding of variation, differentiating between common sources of expected variation inherent to a process (common cause variation) and special sources of variation that can be eliminated \cite{Wheeler_SPC_2010}.  SPC control charts are statistical and visual tools that can be used to easily detect changes in the process \cite{Wheeler_SPC_2010}. Although SPC was originally developed for industrial settings, SPC concepts have also been introduced to healthcare settings. Healthcare applications include process improvement and quality control in surgical care \cite{Woodall,Woodall2,Cheung_Radiographics2012,novoa_Mediastinum_2020,Seim_Anesthesiology2006} and in neonatal intensive care units \cite{Gupta_ClinPerinatol2017}. Recently, Feng et al. \cite{Feng_Nature_2022} discussed the need to adapt quality control methods for the monitoring of clinically-deployed systems. These prior works, which explored the use of SPC for monitoring surgical or intensive care unit activities, motivated us to adapt SPC-based method for monitoring OOD in medical imaging applications.

To successfully implement SPC for process control, it is important to select high-quality monitoring features that represent the task being queried. For instance, our initial investigation showed that using SPC charts to monitor changes in basic image statistics (e.g., mean, standard deviation, skewness, kurtosis) or local texture features, is inadequate for distinguishing chest x-rays from other medical image types (refer to Appendix Figures~\ref{fig:image-statistics} and \ref{fig:texture-features}). Thus, more sophisticated feature extraction techniques are necessary for the successful implementation of an SPC-based framework for OOD detection and input monitoring.

\begin{figure}[!t]
    \centering
    \includegraphics[width=0.95\textwidth]{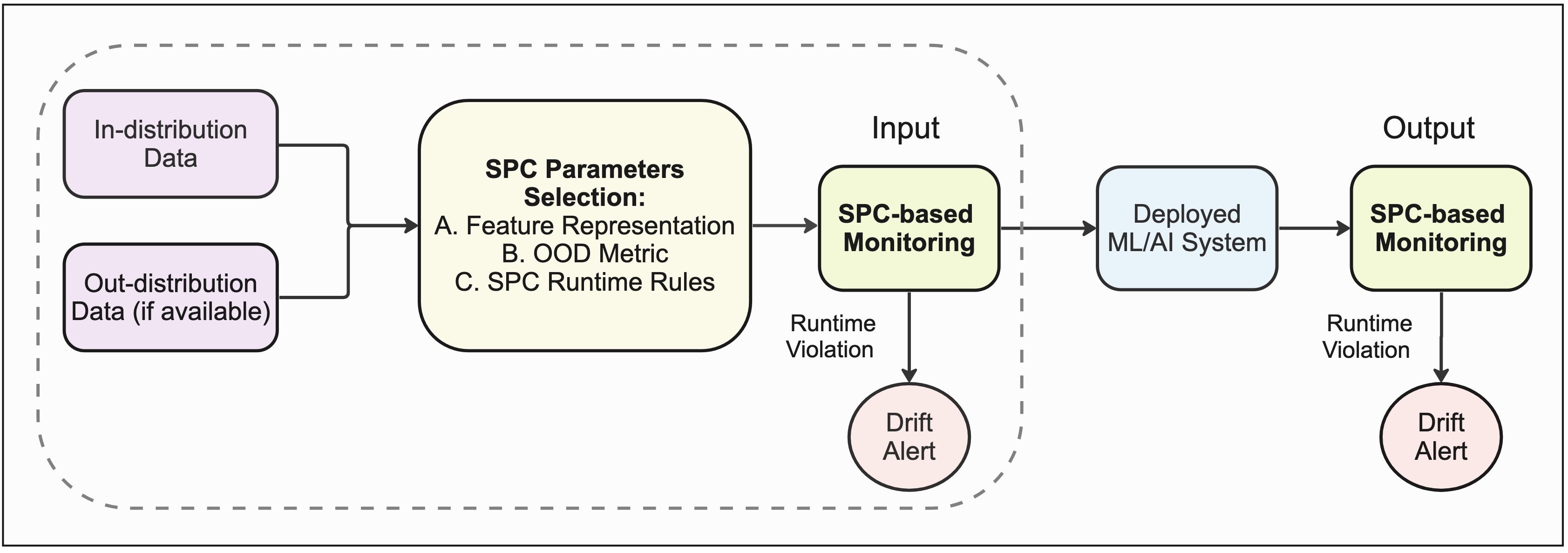}
    \caption{~\textbf{Overview of the proposed SPC-based framework for drift monitoring.} Dashed box indicates that this work focuses on monitoring inputs; however, the same framework can be applied to oversee the model's outputs.}
    \label{fig:Overview}
\end{figure}

Figure~\ref{fig:Overview} illustrates how the proposed framework can be integrated within ML development and how it can be applied for data drift and model's output monitoring post deployment. It is important to emphasize that although this framework is applicable for both input and output monitoring, this paper focuses on input monitoring. As shown in the dashed box in the figure, this framework begins with the selection of a dataset that accurately reflects the expected operating conditions (in-distribution) of the task. Optionally, a representation of the out-distribution data can be provided if such data is known and available. We then use ML methods to extract feature representations of the in-distribution (and if available, out-of-distribution) data. Next, an appropriate distance-based metric for out-distribution detection is chosen; additionally, runtime rules for the SPC charts are established. Finally, the selected SPC-based OOD method is applied to continuously monitor for input data changes as quantified by the OOD metric, signaling any instances that deviate from the expected data distribution.
Depending on the application and observed rate of data drift, corrective actions (e.g., recalibration) may be periodically required to refine the model.

\begin{table}[!t]
\centering
\begin{tabular}{@{}lll@{}}
\toprule
\textbf{Feature Selection} & \textbf{OOD Metrics} & \textbf{SPC Tools/Rules} \\ \midrule
Unsupervised learning & Cosine Similarity & $3\sigma$ SPC \\
Supervised learning & Mahalanobis Distance & CUSUM \\
Supervised contrastive learning & Geodesic Distance  & ADWIN \\
\bottomrule
\end{tabular}
\vspace{5pt}
\caption{~\textbf{SPC parameters selection for designing SPC-based monitoring framework.} The chosen features represent the incoming data stream; the applied OOD metrics quantified changes in the feature space; and SPC tools formalized how OOD detection flags were thrown based on the calculated metrics.}
\label{tab:method_summary}
\end{table}

\subsection{Contributions}
This is the first work that combines SPC with ML methods to provide a practical solution for data drift monitoring in medical imaging. The main contributions of this paper can be summarized as follows. 

\begin{itemize}
    \item We propose a \textbf{new framework for data drift detection and monitoring} that combines machine learning methods for OOD detection with SPC charts. This framework, with appropriate design choices, has been shown to be effective across various imaging modalities, making it modality-agnostic, and capable of detecting data drift, irrespective of the dataset source.

    \item We use the proposed framework for \textbf{OOD detection}. We demonstrate that SPC-based OOD detection requires a careful design and selection of parameters, which are detailed in Table~\ref{tab:method_summary}. Our experiments reveal that the efficacy of the SPC-based method in detecting OOD changes is contingent on the chosen features and their quantification. These insights, accompanied by detailed discussions and results, are presented in Section~\ref{sec:Experiments}, Table~\ref{tab:CT_results_summary}, and Table~\ref{tab:CXR_results_summary}.
    
    \item We use the proposed framework for \textbf{data drift monitoring} in different drift scenarios. We find that different SPC techniques offer unique advantages. Notably, the 3$\sigma$ method is particularly effective for monitoring OOD instances at the individual image level, while Cumulative Sum Control Chart (CUSUM) is adept at quickly identifying persistent subtle drifts in the data distribution. These findings are further elaborated in Section~\ref{Section V}.

\end{itemize}

\vspace{10pt}
\textbf{Outline.} The rest of this paper is organized as follows. Section~\ref{sec:RW} provides a brief review of current OOD methods. In Section~\ref{sec:SectionIII}, we introduce our proposed SPC-based approach for OOD detection and monitoring. Detailed experimental results within the context of two medical imaging tasks are presented in Section~\ref{sec:Experiments}. This is followed by a discussion of these results in Section~\ref{Section V}. The paper concludes with Section~\ref{sec:Conclusion}, summarizing our findings and discussing future directions.

\section{Related Work}
\label{sec:RW}

\subsection{OOD Methods}
Several methods have been proposed, with the main objective to identify data points that deviate from the distribution observed during the training of ML models \cite{yang2021generalized}. This helps models avoid making unreliable predictions and maintain their robustness and reliability.

As described in \cite{sahiner2023data}, data drift refers to changes in the characteristics of the input data fed into ML models. It commonly arises due to variations in patient demographics or data acquisition devices, such as different manufacturers or models of scanners, leading to inconsistencies in the input. Gustav et al. \cite{maartensson2020reliability} explored this type of drift, and found that while different models adapt to brain MR datasets with similar disease populations and protocols as the training data, their performance declines substantially when applied to clinical cohorts showcasing different tissue contrasts. Similarly, De Fauw et al. \cite{de2018clinically} found a degradation in the model performance caused by data drift due to the differences in optical coherence tomography imaging scanners. Another study \cite{magudia2021trials} reported that mislabeling of CT exams during automated data collection is common, which could potentially result in incorrect inputs to AI systems, highlighting a practical challenge in using AI technologies in clinical practice. Other studies  \cite{nestor2019feature,pooch2020can,albadawy2018deep,bernhardt2022potential,finlayson2021clinician,celi2022sources,saez2021potential} investigated data drift originating from distribution changes over time \cite{nestor2019feature}, differences in acquisition and data sources \cite{pooch2020can,albadawy2018deep}, and changes in the patient population \cite{bernhardt2022potential,finlayson2021clinician,celi2022sources}.

Given the serious consequences of data drift in healthcare, several strategies have been proposed to mitigate the effects of such drift \cite{yang2021generalized,sahiner2023data}. During the pre-deployment phase, several strategies can be used including diversifying training data by incorporating samples from multiple sources \cite{sahiner2023data} or integrating synthetic data \cite{badano2018evaluation}. Another pre-deployment strategy is the utilization of the maximum softmax probability in a classification model as the indicator score of in-distribution \cite{hendrycks2016baseline}. Emphasis has also been placed on outlier exposure strategies, which involve exposing the training models to a set of collected OOD samples, or ``outlier'', during training to help models learn in and out discrepancy \cite{papadopoulos2021outlier}. Density-based methods are another OOD strategy that explicitly model the in-distribution with probabilistic models and flag test data in low-density regions as OOD. Finally, distance-based methods have been used to identify OOD samples by measuring how far away the testing samples from the centroids (or other statistics) of in-distribution (training) samples. 

Well-known metrics that have been used to measure this distance include Mahalanobis distance \cite{lee2018simple}, Euclidean distance \cite{huang2020feature}, Geodesic distance \cite{gomes2022igeood}, and cosine similarity \cite{techapanurak2020hyperparameter}. In \cite{lee2018simple}, the authors used the minimum Mahalanobis distance to all class centroids for OOD detection. Other works implemented cosine similarity between test sample features and class features to determine OOD samples \cite{techapanurak2020hyperparameter,gomes2022igeood}. 
It is important to note that non-parametric metrics (e.g., cosine similarity) do not impose distributional assumptions on the feature space, potentially offering more simplicity and flexibility in various contexts as compared to parametric metrics (e.g., Mahalanobis distance). In addition to the metrics, there has been a discussion on the impact of the extracted feature embedding or representation on ODD strategies. For example, it has been found that using the right feature representation can greatly impact the performance of ODD \cite{ming2022cider}, which is consistent with our findings in Section~\ref{sec:Experiments}.

\subsection{OOD Monitoring}
The importance of AI monitoring is becoming increasingly recognized as it offers a direct way to track performance decline after an AI system is deployed \cite{sahiner_BJR_2023}. Monitoring can be conducted either at the input level, focusing on out-of-distribution inputs, or at the output level, examining variations in the difference between the model output and the target variable. While many OOD detection methods have been discussed in the literature, most are tailored for the pre-deployment phase, focus on the output, and often overlook the temporal aspect. Only a handful of studies \cite{Feng_Nature_2022,soin2022chexstray,sahiner_BJR_2023} discussed integrating quality improvement and temporal techniques into AI performance tracking. However, these studies mainly provide a theoretical foundation, and do not discuss practical implementations. In this study, we introduce the concepts of SPC for AI monitoring by presenting a practical implementation of SPC charts for OOD detection and data drift monitoring. This includes a detailed exploration of essential factors such as feature selection, metrics, and various monitoring options.

\section{Methods}
\label{sec:SectionIII}
In this section, we present our framework for detecting individual OOD images and monitoring data drift, which involves the following steps: (1) training and fine-tuning a feature extraction model appropriate for a given task ($1^{st}$ column in Table~\ref{tab:method_summary}); (2) choosing the OOD metric that quantifies the feature space and separates OOD ($2^{nd}$ column in Table~\ref{tab:method_summary}); and (3) generating SPC charts using the computed OOD metrics ($3^{rd}$ column in Table~\ref{tab:method_summary}). Each of these steps is described next.

\subsection{Feature Representation}
\label{sec:Feature}
To determine the best feature representation, we investigated different approaches for training a feature extraction model using available in-distribution and out-of-distribution (OOD) data. This included both unsupervised and supervised deep features, as described below.

Before proceeding, we note that we initially established that zero-order image statistics and texture features would be insufficient to detect OODs (see Figures~\ref{fig:image-statistics} and \ref{fig:texture-features} in the Appendix). This baseline underscored the need to use more complex features derived from deep learning.

\subsubsection{Autoencoder Architecture (Unsupervised Learning)} For the unsupervised learning method, we used an autoencoder to extract feature representation from the data. The model's architecture included $16$ hidden channels, and a latent dimension of $100$ to provide a balance between complexity and computational efficiency. The encoder was composed of $5$ convolutional layers followed by $5$ ReLU activation layers, while the decoder mirrored this structure with $5$ convolutional transpose layers and $5$ ReLU layers. The autoencoder is configured with a specific set of parameters to optimize its performance for a specific task (Table \ref{tab:autoencoder}).

\subsubsection{Convolutional Neural Networks (Supervised Learning)}
To extract relevant feature representations using supervised learning, we trained a convolutional neural network (CNN) with two methods: binary cross-entropy (BCE) and contrastive learning. We chose contrastive learning as a training method for its proven effectiveness in learning discriminative features \cite{le2020contrastive}. This method focuses on identifying and capturing essential characteristics and similarities within the data, making it highly effective for differentiating between various classes. After training task-specific models using binary cross-entropy and contrastive learning, we extract deep features from either the logits layer (in the CT task) or the final convolutional layer before the fully connected layer (CXR task). This supervised learning approach for extracting feature representation requires task-specific labels for training. 

\vspace{5pt}
These extracted features are then utilized to calculate distance-based OOD metrics.

\subsection{OOD Metrics}
\label{Sec:SPCmetrics}
To measure the distance between out-distribution samples and in-distribution samples, we used two geometric-based metrics: 1) Mahalanobis distance (MD) and 2) cosine similarity (CS). Although less explored, distance-based methods can be applied to wide range of architectures due to their operation within the representation space. This capability enables them to measure the distance between out-of-distribution and in-distribution data effectively, bypassing the need for model-specific adjustments (i.e., model-agnostic). 

\subsubsection{Mahalanobis distance}
The Mahalanobis distance is a parametric metric, which calculates the distance of a point from a distribution by considering the mean and the covariance matrix. It estimates the probability density based on measurements from Gaussian distributed data, which is a reasonable data distribution assumption for the large \cite{Park_Neurips2021}, multi-dimension feature spaces explored here. The formula \cite{Park_Neurips2021} for the Mahalanobis 
distance \( D_M \) of a point \( \mathbf{x} \) from a group of points with mean \( \boldsymbol{\mu} \) and covariance matrix \( \mathbf{S} \) is given by:

\[
D_M(\mathbf{x}) = \sqrt{(\mathbf{x} - \boldsymbol{\mu})^\top \mathbf{S}^{-1} (\mathbf{x} - \boldsymbol{\mu})}
\]

Where \( \mathbf{x} \) is the vector of the point whose distance is being measured, \( \boldsymbol{\mu} \) is the mean vector of the points in the distribution, \( \mathbf{S} \) is the covariance matrix of the distribution, \( \mathbf{S}^{-1} \) is the inverse of the covariance matrix, and \( (\mathbf{x} - \boldsymbol{\mu})^\top \) is the transpose of the difference between the point vector and the mean vector. The Mahalanobis distance accounts for the variance of each component (or feature) of the points and the covariance between components, making it a more robust distance metric in multivariate spaces compared to the Euclidean distance.

\subsubsection{Cosine Similarity}
Cosine similarity is a non-parametric measure that evaluates the similarity between two vectors by calculating the cosine of the angle between them. Mathematically, cosine similarity between two 
vectors \( \mathbf{A} \) and \( \mathbf{B} \) is formulated as follows \cite{Park_Neurips2021}:

\[
\text{Cosine Sim}(\mathbf{A}, \mathbf{B}) = \frac{\mathbf{A} \cdot \mathbf{B}}{\|\mathbf{A}\| \|\mathbf{B}\|}
\]

Where \( \mathbf{A} \cdot \mathbf{B} \) represents the dot product of vectors \( \mathbf{A} \) and \( \mathbf{B} \), and \( \|\mathbf{A}\| \) and \( \|\mathbf{B}\| \) are the magnitudes (or norms) of vectors \( \mathbf{A} \) and \( \mathbf{B} \), respectively. Cosine similarity ranges from -1 to 1. A value of 1 implies that the vectors are identical in orientation, 0 indicates orthogonality (no similarity), and -1 signifies that the vectors are diametrically opposed.

 \vspace{5pt}
By utilizing these geometric distances, we were able to quantify the similarity between in-distribution and out-distribution feature vectors. We then use the computed OOD distance as the input for SPC charts. 

\subsection{SPC for OOD Detection and Monitoring}
\label{Sec:SPCmethods}

In this study, we used two types of SPC charts for OOD detection and monitoring: $3\sigma$ control charts and Cumulative Sum Control Charts (CUSUM). Our selection for these two methods was informed by our review of the literature and empirical evidence supporting their effectiveness in various domains \cite{Wheeler_SPC_2010}. 

The following subsections detail how we used the proposed method to detect OOD images, and how we created simulations of data drift to assess the effectiveness of the proposed method for OOD monitoring, utilizing both $3\sigma$ and CUSUM charts.

\subsubsection{Detecting OOD Images}
For each method of feature extraction and scoring, we calculated the mean $\mu$ and standard deviation $\sigma$ of the metric used for OOD detection (CS or MD) for the in-distribution training data. These metric statistics, $\mu$ and $\sigma$, establish the control limits for SPC analysis. We then computed a mean vector representing the average values of each feature across all in-distribution images. This mean vector is then used as a reference point to evaluate the distance metrics (OOD metrics), by comparing the feature vector of each test image (which could originate from either in-distribution or out-of-distribution data) against it. Moreover, for the Mahalanobis distance calculation, we also derived the covariance matrix from the in-distribution data. Ultimately, a test image is identified as either in-distribution or out-of-distribution based on whether its computed OOD metric (CS or MD) falls outside the control limits defined by the \(\mu\) and \(\sigma\) of the training dataset.

\subsubsection{Temporal Simulation of Data Drift} \label{sec:Sim}
To assess the effectiveness of our SPC-based method for OOD monitoring, we created a simulation scenario spanning a time series from day $1$ to day $N$, where each day comprises a batch of $d$ images. We then computed a daily metric by averaging the OOD metrics (e.g., cosine similarity) across all images within each batch, thus representing the data of each day as a singular metric. Midway through the simulation—at day $N/2$—we introduced data drift by increasing the daily percentage of OOD inputs. This simulation is designed to evaluate the adaptability of our method to data pattern changes, mirroring real-world conditions where AI models might face variations from their initial training data. For the monitoring and identification of OOD data, we used the day-averaged $3\sigma$ method and CUSUM.

\subsubsection{Three Sigma Control Chart}
The three sigma ($3\sigma$) control chart is a widely-used method where a process is considered out-of-control if the process outputs deviate beyond the range of $[\mu - 3\sigma, \mu + 3\sigma]$, where $\mu$ denotes the mean and $\sigma$ denotes the standard deviation. In our study, we flag an individual test image or a day's batch of test images as out-of-distribution if their metric exceeds this $3\sigma$ range, determined by the mean and standard deviation of in-distribution images. Although we did not explore additional control limits in this work, it is important to note that values of 1 or 2$\sigma$ can also be employed to accommodate the specific needs and sensitivities of different applications.

\subsubsection{Cumulative Sum Control Chart (CUSUM)}
CUSUM is a type of SPC chart utilized to monitor small shifts or changes in a process over time \cite{Page-CUSUM}. CUSUM plots the cumulative sum of deviations of each sample value from a target value, effectively highlighting trends or shifts. CUSUM calculates the cumulative sum of differences between individual data points and a target value or mean, expressed in its simplest form as a pair of equations \cite{crosier1986new}. The upper control limit

\begin{equation}\label{eq:CUSUM1}
S_i^+ = \max ( 0, S_{i-1}^+ + (x_i - \mu_0 - k))
\end{equation}

and the lower control limit

\begin{equation}\label{eq:CUSUM2}
S_i^- = \max ( 0, S_{i-1}^- - (x_i - \mu_0 + k))
\end{equation}
are the cumulative sums for detecting upward and downward shifts, respectively. Here, \( x_i \) is the \( i \)-th individual data point, \( \mu_0 \) is the mean of the in-distribution process, and \( k \) is an allowance, often set as \( \frac{\sigma}{2} \). An alarm is triggered when either $S_i^+ > h$ or $S_i^- > h$, where \( h \) is the decision interval or threshold, commonly set to \( 4 \times \sigma \). This decision rule implies that if either \( S_i^+ \) or \( S_i^- \) exceeds \( h \), a potential shift in the process has occurred, necessitating further investigation or action. The sensitivity of CUSUM is adjustable by modifying the allowance \( k \) and the threshold \( h \). Balancing CUSUM's sensitivity involves a trade-off between false detection (type I errors) and detection delays or failures (type II errors). Therefore, setting CUSUM parameters involves striking a balance between minimizing false detection and accepting a tolerable delay in detection.

\section{Experiments and Results}
\label{sec:Experiments}
We evaluated the proposed SPC-based method for both identifying individual OOD inputs and demonstrating the ability to detect shifts in data over time on the following tasks: 1) differentiation of axial vs. non-axial CT images and 2) differentiation of CXR vs. other radiography images, and 3) differentiation of adult CXR vs. pediatric CXR. For these tasks, we illustrated the trade-offs inherent to different feature extraction and distance metric choices.

Each task is structured into a four-pronged description of: 1) the used dataset, 2) the process of selecting SPC parameters, 3) OOD image detection results, and 4) OOD monitoring results. In the monitoring experiments for both CT and CXR tasks, it was assumed that the duration spanned 60 days ($N=60$) and that each day involved the analysis of 100 images ($d=100$) per day.

\begin{figure}[!h]
     \centering
     \begin{subfigure}[b]{\linewidth}
     \centering
         \includegraphics[width=0.65\linewidth]{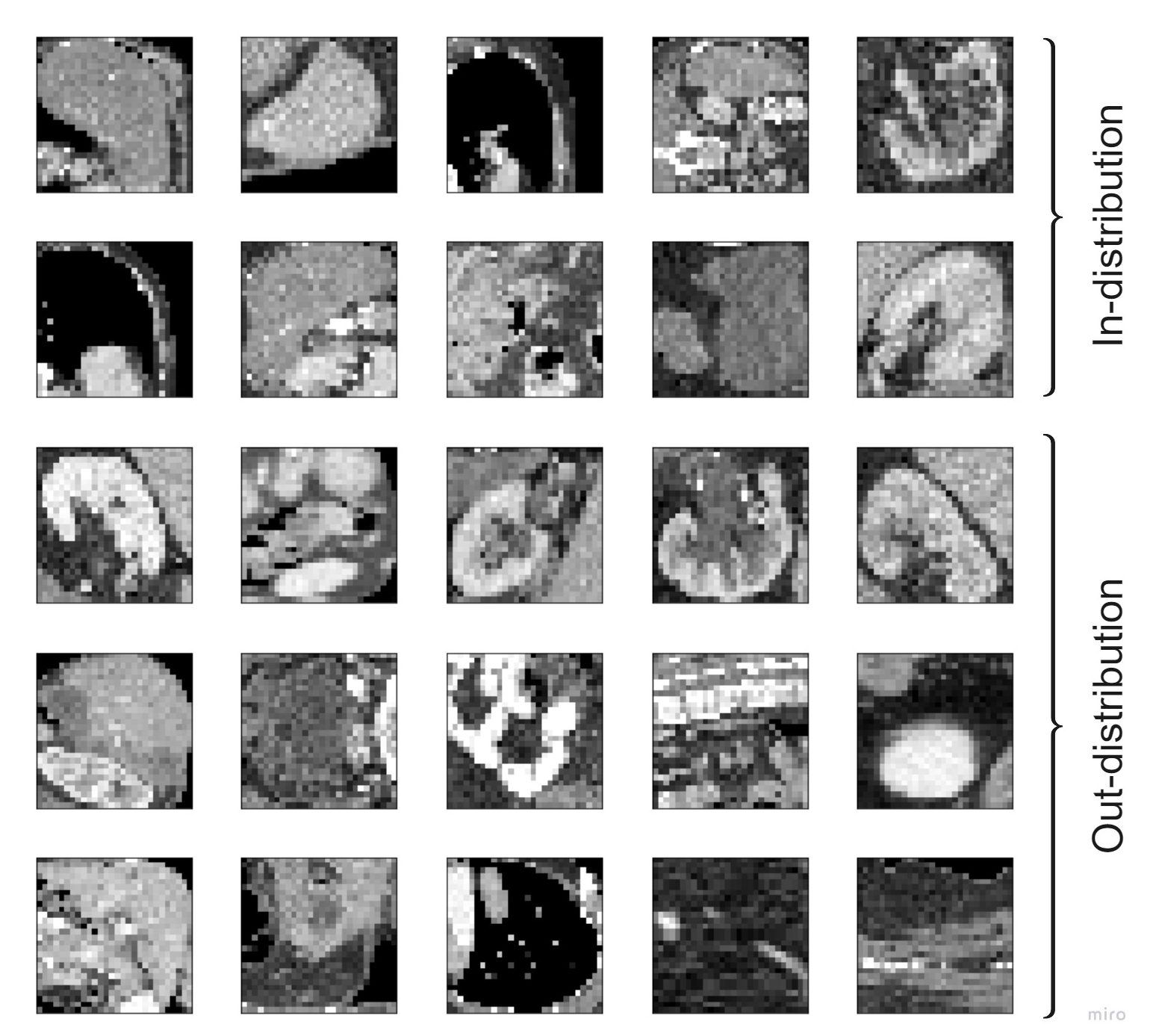}
         \caption{~Examples of axial CT images (in-distribution) and coronal/sagittal CT images (out-of-distribution).\\}
         \label{fig:medmnist_images}
     \end{subfigure}
     \begin{subfigure}[b]{\linewidth}
     \centering
         \includegraphics[width=0.65\linewidth]{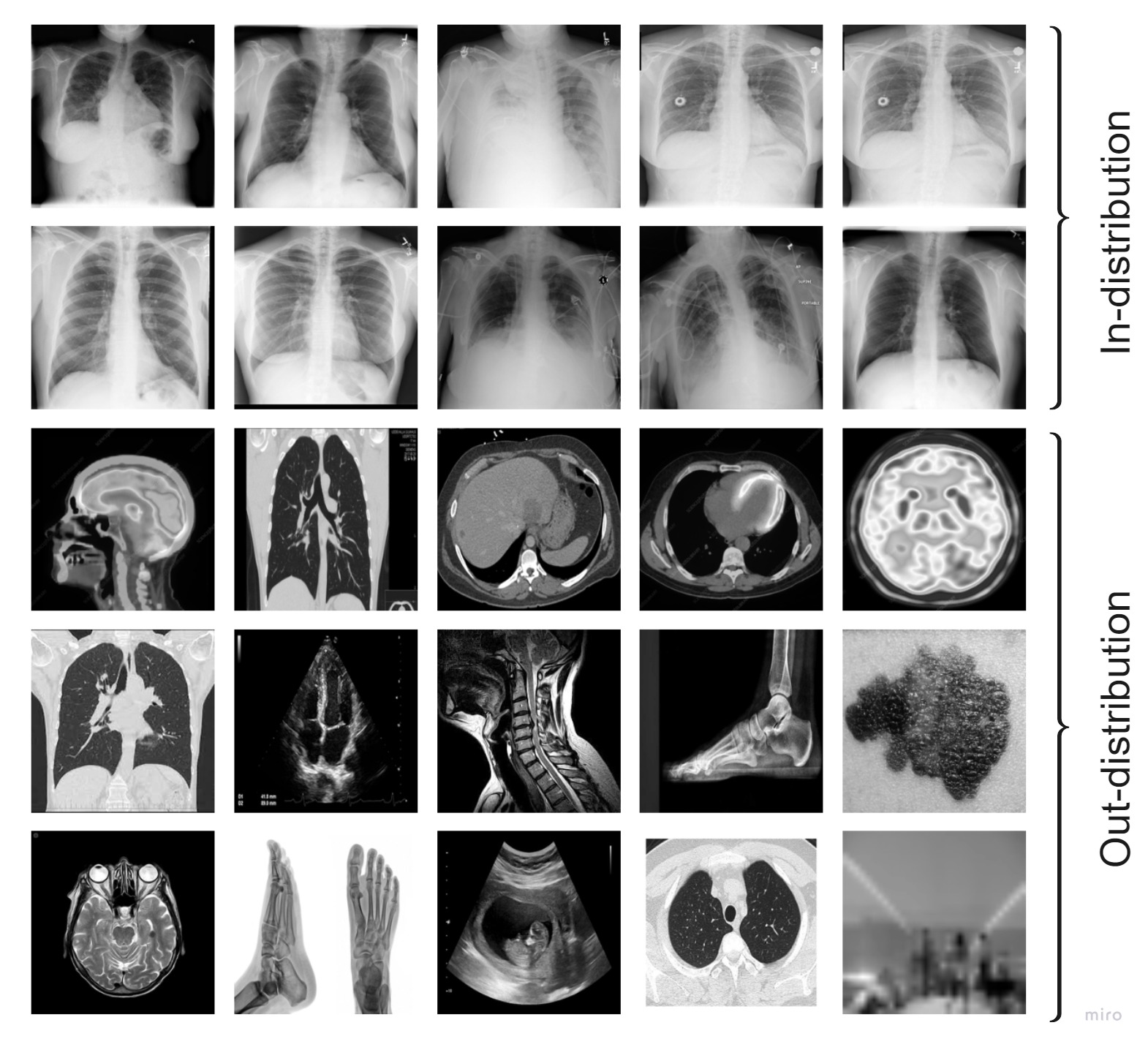}
         \caption{~Examples of CXR images (in-distribution) and non-CXR images (out-distribution).\\}
         \label{fig:cxr_images}
     \end{subfigure}
     \caption{~ \textbf{Example of in-distribution and out-of-distribution images from open-source datasets used in the experiments.}}
 \end{figure}

\subsection{Detecting and Monitoring OOD CT Scans} \label{sec:Experiments.CT}
The OOD CT task is designed to distinguish between axial CT images (in-distribution) and coronal/sagittal CT images (out-of-distribution, OOD). This task is motivated by the fact that ML models in CT analysis are typically trained on specific image orientations; therefore, multiplanar reformatted images, which deviate from the training distribution, might lead to reduced performance.

\subsubsection{Dataset}\label{sec:data.CT}
We used CT images available from the open-source Medical MNIST dataset \cite{yang_medmnist_2023}. Specifically, we used OrganMNIST subset of the Medical MNIST dataset; the dataset authors sliced 3D CT scans (from the Liver Tumor Segmentation Benchmark (LiTS) \cite{bilic_liver_2023} along three different viewing planes -- axial, coronal, and sagittal -- and pre-processed them into $28\times28$ grayscale images, similar to digit MNIST. All the images were normalized ($\mu=0.5$, $\sigma=0.5$) prior to model training. Figure~\ref{fig:medmnist_images} shows examples of in-distribution (axial) and out-of-distribution (coronal/sagittal) images for this dataset. We note that axial CT images have different views than coronal and sagittal CT images. Axial CT captures cross-sectional views of the body horizontally from top to bottom, while coronal images slice it vertically from front to back and sagittal images divide it into left and right halves, offering distinct perspectives of internal structures.

\subsubsection{Model Specifications, Training Details, and Feature Extraction}\label{sec:CT_Exp_details}
We compared the three feature extraction methods: 1) an unsupervised convolutional autoencoder, 2) a slice-supervised binary cross-entropy (BCE) ResNet-18, and 3) slice-supervised contrastive ResNet-18 (Section \ref{sec:Feature}). All these models were trained using $57.11\%$ and validated using $10.52\%$ of the available CT images in the OrganMNIST subset. These training and validation splits were created by the MedMNIST authors. We refer the reader to Tables \ref{tab:autoencoder}, \ref{tab:supervised_BCE}, and \ref{tab:supervised_contrastive} in the Appendix for more details about the hyper-parameter settings and selection for all three methods.

\subsubsection{Results for SPC-based OOD Detection}
Table~\ref{tab:CT_results_summary} reports the image-level OOD detection performance across the different feature extraction methods and distance metrics. We calculated the accuracy, sensitivity, and specificity with bootstrapped confidence intervals ($n=100$) over random sub-samples ($\text{size} = 500$) of the test set. Images were flagged if their OOD metrics fell outside the $\mu \pm 3\sigma$ control limits. While the autoencoder resulted in the smallest mean accuracy (CS: $0.509$; MD: $0.511$) and mean sensitivity (CS: $0.014$; MD: $0.003$), it yielded the highest mean specificity (CS: $0.989$; MD: $0.999$). The low sensitivity and high specificity for the autoencoder could be explained by its feature representation: incoming images all had features with similar distances from the training distribution, making OODs indistinguishable for $3\sigma$ flagging. Thus, the model flagged only a few samples as OOD, yielding the reported low sensitivity and high specificity. 

Slice supervision, meaning each CT slice was labeled as in-distribution or out-of-distribution at training time, significantly improved accuracy and sensitivity compared to the unsupervised baseline ($p \ll 0.01$ for all comparisons). Among the supervised methods, contrastive ResNet-18 features with cosine similarity resulted in the highest mean accuracy (CS: $0.913$) and specificity (CS: $0.848$). Mean sensitivity was similar between contrastive (CS: $0.980$) and BCE ResNet-18 (CS: $0.985$) features. Figure \ref{fig:CTa} visualizes the image-level $3\sigma$ cosine similarity detection of contrastive ResNet-18 features. The figure illustrates that most OOD images are identified with a high degree of confidence, evidenced by their large difference compared to the mean cosine similarity. Therefore, given the favorable sensitivity-specificity tradeoff and robust in- vs. out-distribution separation in feature space, we used slice-supervised ResNet-18 features and the cosine similarity metric for the subsequent OOD monitoring experiment.

\subsubsection{Results for SPC-based OOD Monitoring}
\begin{figure}[!t]
     \begin{subfigure}[b]{\linewidth}
        \centering

         \includegraphics[width=0.70\linewidth]{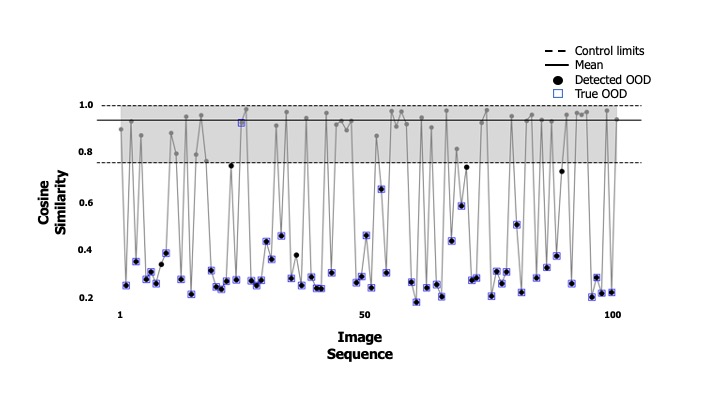}
         \caption{$3\sigma$ rule for OOD detection of individual CT images. The figure is visualized for a randomly chosen subset of 100 images from the test set. Each point represents a single image. Blue squares highlight ground truth OOD images; black circles are flagged images due to scores outside the limits.\\}
         \label{fig:CTa}
     \end{subfigure}
     \begin{subfigure}[b]{\linewidth}
     \centering
         \includegraphics[width=0.70\linewidth]{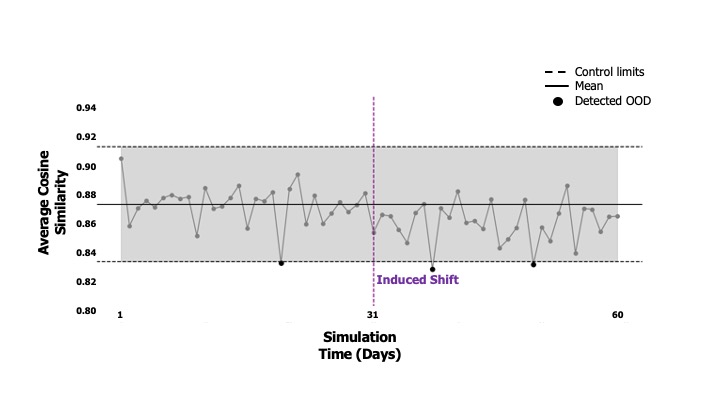}
         \caption{$3\sigma$ rule applied on daily averaged scores to detect data drift, i.e., increase in non-axial (OOD) inputs.\\}
         \label{fig:CTb}
     \end{subfigure}
     \centering
     \begin{subfigure}[b]{\linewidth}
          \centering

         \includegraphics[width=0.70\linewidth]{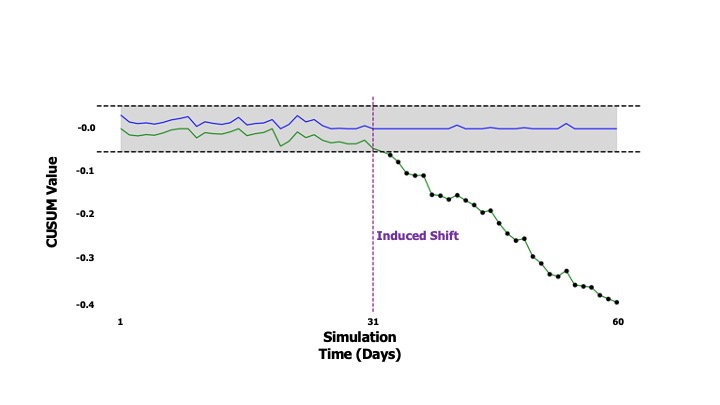}
         \caption{CUSUM monitoring ($k=0.10$) to detect data drift, i.e., increase in non-axial (OOD) inputs. The blue and green signals represent the high- and low-side CUSUM signals, respectively. Black dashed lines indicate $\pm$ thresholds. Black dots are OOD flags. Purple dashed line indicates induced shift.\\}
         \label{fig:CTc}
     \end{subfigure}
     \caption{~ \textbf{SPC-based detection of OOD non-axial CT slices and monitoring for data-drift.} 
     }
     \label{fig:CT_CUSUM}
 \end{figure}

In our CT monitoring simulation (see Section~\ref{sec:Sim}), we assumed a clinical facility collects and labels 100 abdominal CT slices per day. During the first month, we set the daily OOD percentage uniformly between 0-1\%. In the second month, we increased the OOD rate to be uniformly sampled from 3-5\%. Figures \ref{fig:CTb} and \ref{fig:CTc} highlight the difference in OOD sensitivity between $3\sigma$ and CUSUM. The $3\sigma$ monitoring throws a flag on day 37; however, subsequent days are well-within the control limits, suggesting that the flag might have been an anomaly. On the other hand, CUSUM monitoring reports a flag two days after the induced shift. The low-side CUSUM signal significantly decreases after this day, lending support to the initial flag. The parameter $k$ can also be tuned to reduce detection delay (see Table \ref{tab:tuningk} in the Appendix).

\vspace{5pt}
These results suggest that in this batched longitudinal setting, CUSUM is a more robust feature monitoring scheme, given well-formed features and distance metrics. Additionally, our experiments suggest that the features derived from contrastive learning, paired with cosine similarity, yield effective results for the CT application. This highlights the significance of selecting appropriate features and metrics when employing SPC-based techniques for OOD detection and monitoring.

\begin{table*}[!t]
\centering
\scriptsize
\begin{tabular}{|c|ccc|ccc|ccc|}
\hline
\multirow{2}{*}{\textbf{OOD Metric}} & \multicolumn{3}{c|}{\textbf{Unsupervised Convolutional Autoencoder}} & \multicolumn{3}{c|}{\textbf{Supervised BCE ResNet-18}} & \multicolumn{3}{c|}{\textbf{Supervised Contrastive ResNet-18}} \\
\cline{2-10} & \multicolumn{1}{c|}{Accuracy} & \multicolumn{1}{c|}{Sensitivity} & Specificity & \multicolumn{1}{c|}{Accuracy} & \multicolumn{1}{c|}{Sensitivity} & Specificity & \multicolumn{1}{c|}{Accuracy} & \multicolumn{1}{c|}{Sensitivity} & Specificity \\
\hline
Cosine  &
\multicolumn{1}{c|}{\begin{tabular}[c]{@{}c@{}}0.509 \\ {[}0.481-0.548{]}\end{tabular}} &
\multicolumn{1}{c|}{\begin{tabular}[c]{@{}c@{}}0.014 \\ {[}0.002-0.029{]}\end{tabular}} &
\multicolumn{1}{c|}{\begin{tabular}[c]{@{}c@{}}0.989 \\ {[}0.976-1.000{]}\end{tabular}} &
\multicolumn{1}{c|}{\begin{tabular}[c]{@{}c@{}}0.906 \\ {[}0.88-0.93{]}\end{tabular}} &
\multicolumn{1}{c|}{\begin{tabular}[c]{@{}c@{}}\textbf{0.985} \\ {[}0.97-0.99{]}\end{tabular}} &
\multicolumn{1}{c|}{\begin{tabular}[c]{@{}c@{}}0.830 \\ {[}0.789-0.870{]}\end{tabular}} &
\multicolumn{1}{c|}{\begin{tabular}[c]{@{}c@{}}\textbf{0.913} \\ {[}0.89-0.93{]}\end{tabular}} &
\multicolumn{1}{c|}{\begin{tabular}[c]{@{}c@{}}0.980 \\ {[}0.96-0.99{]}\end{tabular}} &
\multicolumn{1}{c|}{\begin{tabular}[c]{@{}c@{}}0.848 \\ {[}0.80-0.88{]}\end{tabular}} \\
\hline
Mahalanobis &
\multicolumn{1}{c|}{\begin{tabular}[c]{@{}c@{}}0.511 \\ {[}0.48-0.56{]}\end{tabular}} &
\multicolumn{1}{c|}{\begin{tabular}[c]{@{}c@{}}0.003 \\ {[}0.00-0.01{]}\end{tabular}} &
\multicolumn{1}{c|}{\begin{tabular}[c]{@{}c@{}}\textbf{0.999} \\ {[}0.99-1.00{]}\end{tabular}} &
\multicolumn{1}{c|}{\begin{tabular}[c]{@{}c@{}}0.911 \\ {[}0.89-0.93{]}\end{tabular}} &
\multicolumn{1}{c|}{\begin{tabular}[c]{@{}c@{}}0.977 \\ {[}0.96-0.99{]}\end{tabular}} &
\multicolumn{1}{c|}{\begin{tabular}[c]{@{}c@{}}0.849 \\ {[}0.81-0.89{]}\end{tabular}} &
\multicolumn{1}{c|}{\begin{tabular}[c]{@{}c@{}}0.906 \\ {[}0.88-0.93{]}\end{tabular}} &
\multicolumn{1}{c|}{\begin{tabular}[c]{@{}c@{}}0.966 \\ {[}0.94-0.99{]}\end{tabular}} &
\multicolumn{1}{c|}{\begin{tabular}[c]{@{}c@{}}0.849 \\ {[}0.81-0.89{]}\end{tabular}} \\
\hline
\end{tabular}
\\
\caption{~\textbf{Performance of $3\sigma$ OOD detection of non-axial CT slices for each feature representation and distance metric.} $95\%$ confidence intervals were calculated by bootstrapping with $n=100$ samples over test-set subsets of size $m=500$. The best performance is highlighted in bold.}
\label{tab:CT_results_summary}
\end{table*}

\begin{table*}[!t]
\centering
\scriptsize
\begin{tabular}{|c|ccc|ccc|ccc|}
\hline
\multirow{2}{*}{\textbf{OOD Metric}} & \multicolumn{3}{c|}{\textbf{Unsupervised Convolutional Autoencoder}} & \multicolumn{3}{c|}{\textbf{Supervised BCE VGG16}} & \multicolumn{3}{c|}{\textbf{Supervised Contrastive VGG16}} \\
\cline{2-10} & \multicolumn{1}{c|}{Accuracy} & \multicolumn{1}{c|}{Sensitivity} & Specificity & \multicolumn{1}{c|}{Accuracy} & \multicolumn{1}{c|}{Sensitivity} & Specificity & \multicolumn{1}{c|}{Accuracy} & \multicolumn{1}{c|}{Sensitivity} & Specificity \\

\hline
Cosine  &
\multicolumn{1}{c|}{\begin{tabular}[c]{@{}c@{}}{0.933} \\ {[}0.91-0.95{]}\end{tabular}} &
\multicolumn{1}{c|}{\begin{tabular}[c]{@{}c@{}}{0.921} \\ {[}0.89-0.94{]}\end{tabular}} &
\multicolumn{1}{c|}{\begin{tabular}[c]{@{}c@{}}{0.944} \\ {[}0.92-0.96{]}\end{tabular}} &
\multicolumn{1}{c|}{\begin{tabular}[c]{@{}c@{}}\textbf{0.95} \\ {[}0.98-0.96{]}\end{tabular}} &
\multicolumn{1}{c|}{\begin{tabular}[c]{@{}c@{}}\textbf{0.98} \\ {[}0.97-0.98{]}\end{tabular}} &
\multicolumn{1}{c|}{\begin{tabular}[c]{@{}c@{}}\textbf{0.991} \\ {[}0.98-0.99{]}\end{tabular}} &
\multicolumn{1}{c|}{\begin{tabular}[c]{@{}c@{}}0.959 \\ {[}0.95-0.96{]}\end{tabular}} &
\multicolumn{1}{c|}{\begin{tabular}[c]{@{}c@{}} 0.932 \\ {[}0.93-0.95{]} \end{tabular}} &
\multicolumn{1}{c|}{\begin{tabular}[c]{@{}c@{}} 0.948 \\ {[}0.95-0.97{]} \end{tabular}} \\
\hline
Mahalanobis &
\multicolumn{1}{c|}{\begin{tabular}[c]{@{}c@{}}{0.912} \\ {[}0.90-0.93{]}\end{tabular}} &
\multicolumn{1}{c|}{\begin{tabular}[c]{@{}c@{}}{0.905} \\ {[}0.87-0.92{]}\end{tabular}} &
\multicolumn{1}{c|}{\begin{tabular}[c]{@{}c@{}}{0.951} \\ {[}0.91-0.96{]}\end{tabular}} &
\multicolumn{1}{c|}{\begin{tabular}[c]{@{}c@{}}{0.984} \\ {[}0.96-0.98{]}\end{tabular}} &
\multicolumn{1}{c|}{\begin{tabular}[c]{@{}c@{}}{0.971} \\ {[}0.96-0.98{]}\end{tabular}} &
\multicolumn{1}{c|}{\begin{tabular}[c]{@{}c@{}}{0.985} \\ {[}0.97-0.99{]}\end{tabular}} & 
\multicolumn{1}{c|}{\begin{tabular}[c]{@{}c@{}} 0.941 \\ {[}0.93-0.94{]}\end{tabular}} &

\multicolumn{1}{c|}{\begin{tabular}[c]{@{}c@{}}0.929 \\ {[}0.93-0.93{]}\end{tabular}} &
\multicolumn{1}{c|}{\begin{tabular}[c]{@{}c@{}}0.943 \\ {[}0.93-0.96{]}\end{tabular}} \\
\hline 
\end{tabular}
\\
\vspace{7pt}
\caption{~\textbf{Performance of $3\sigma$ OOD detection of non-CXR images for each feature representation and distance metric.} $95\%$ confidence intervals were calculated by bootstrapping with $n=100$ samples over test-set subsets of size $m=500$. The best performance is highlighted in bold.}
\label{tab:CXR_results_summary}
\end{table*}

\subsection{Detecting and Monitoring OOD CXR Images} \label{sec:Experiments.CXR}
Here, the task focuses on distinguishing between CXR images and non-CXR images as well as adult CXR images from pediatric CXR images. This task operates under the assumption that developers do not have knowledge (\textit{a priori}) about the potential characteristics of OOD images; their familiarity is limited to the standard CXR images utilized during the model's training phase.

\subsubsection{Dataset}\label{sec:data.CXR}
We used the NIH CXR dataset \cite{wang_chestx-ray8_2017} as our training dataset to define in-distribution images. The NIH CXR dataset contains 112,120 frontal-view X-ray images from 30,805 distinct patients, and fourteen disease labels extracted from corresponding radiological reports. For this task, we assumed that out-distribution data is unknown, and thus we approached feature extraction by training a model trained to classify CXR images as either healthy or abnormal. However, it is important to note that although we used a binary classification model, other models (e.g., multi-label or pneumonia detection), could also be utilized effectively for learning and extracting in-distribution feature representations without knowing the nature of OOD data.

We used two additional datasets to represent OOD inputs. The first testing set contains non-CXR radiological images such as CT images and bone X-ray images (see samples in Figure~\ref{fig:cxr_images}). The inclusion of other imaging modalities enabled us to test the ability of the method to detect out-of-modality data. The second test set contained pediatric CXRs from the open-source Pediatric Pneumonia Chest X-ray dataset \cite{kermany2018identifying}. For examples of pediatric CXR images, refer to Figure~\ref{fig:PedCXR_Example} in the Appendix. This pediatric dataset enables the assessment of how well our approach, trained on adult CXR, can identify demographic changes and flag the pediatric set as OOD. These test sets enable us to investigate two clinically relevant scenarios: out-of-modality, which involves identifying images from an incorrect modality, and within-modality, focusing on discerning differences in key patient demographic changes within the same modality. Both scenarios are currently handled by human readers in radiological workflows.

All images from the training and testing sets were resized to $256 \times 256$ to meet the requirements of the model (VGG16). 

\subsubsection{Model Specifications, Training Details, and Feature Extraction}\label{sec:CXR_Exp_details}
We compared three feature extraction methods (Section \ref{sec:Feature}): 1) an unsupervised convolutional autoencoder, 2) a supervised binary cross-entropy (BCE) VGG16, and 3) supervised contrastive VGG16. Models were trained using disease labels (i.e., healthy vs. abnormal) readily available in the dataset. Our hypothesis was that, in the absence of direct OOD supervision, the use of disease labels would improve the grouping of chest CXR-derived features in comparison to OOD images such as non-CXR. We trained models using 80\% and validated them using 20\% of available CXRs in the NIH CXR dataset. Details about the models' hyper-parameters can be found in Appendix tables \ref{tab:autoencoder}, \ref{tab:supervised_BCE}, and \ref{tab:supervised_contrastive}.

\subsubsection{Results for SPC-based OOD Detection}

\begin{figure}[!t]
     \centering
     \begin{subfigure}[b]{\linewidth}
          \centering

         \includegraphics[width=0.70\linewidth]{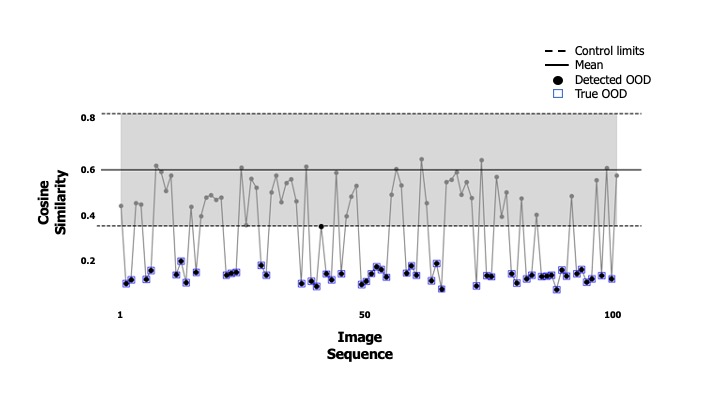}
         \caption{$3\sigma$ rule for OOD detection of individual non-CXR images. The figure is visualized for a randomly chosen subset of 100 images from the test set. Each point represents a single image. Blue squares highlight ground truth OOD images; black circles are flagged images due to scores outside the limits.\\}
         \label{fig:CXRa}
     \end{subfigure}
     \hfill
     \begin{subfigure}[b]{\linewidth}
          \centering

         \includegraphics[width=0.70\linewidth]{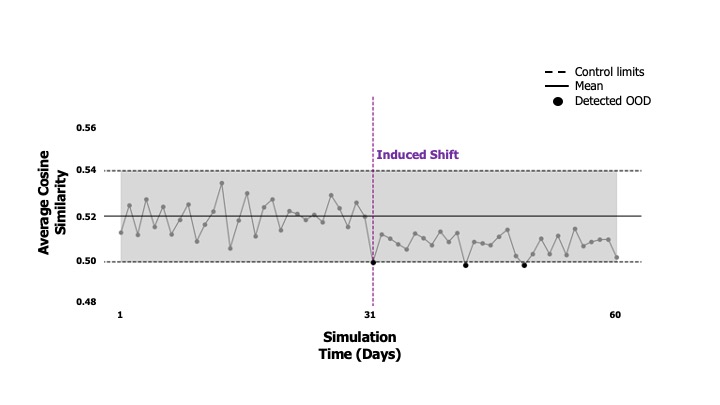}
         \caption{$3\sigma$ rule applied on daily averaged scores to detect data drift, i.e., increase in non-CXR (OOD) inputs.\\}
         \label{fig:CXRb}
     \end{subfigure}
  
     \hfill
     \begin{subfigure}[b]{\linewidth}
          \centering

         \includegraphics[width=0.70\linewidth]{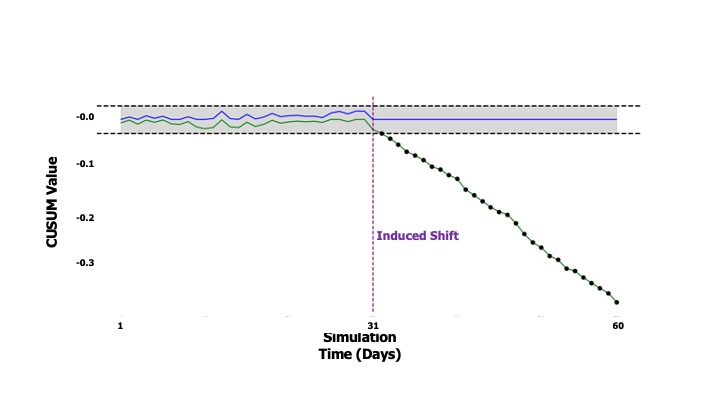}
         \caption{CUSUM monitoring ($k=0.50$) to detect data drift, i.e., increase in non-CXR (OOD) inputs. The blue and green signals represent the high- and low-side CUSUM signals, respectively. Black dashed lines indicate $\pm$ thresholds. Black dots are OOD flags. Purple dashed line indicates the day the shift is induced.\\}
         \label{fig:CXRc}
     \end{subfigure}
     \caption{~\textbf{SPC-based detection of OOD non-CXR images and monitoring for data-drift.}} 
     
     \label{fig:CXR_CUSUM}
 \end{figure}

Table~\ref{tab:CXR_results_summary} reports OOD detection performance on a CXR task. Although we observed variations in performance across the feature extraction methods and distance metrics, all methods achieved acceptable performance. Notably, the supervised VGG16 demonstrated superior accuracy (CS: $0.995$) and sensitivity (CS: $0.984$) compared to the unsupervised baseline, with all values reported within a 100-sample bootstrapped 95\% confidence interval. Interestingly, unlike in CT task (Table~\ref{tab:CT_results_summary}), the supervised BCE-loss VGG16 slightly outperformed the model trained with contrastive-loss. In all results within Table~\ref{tab:CXR_results_summary}, feature quantification using cosine similarity generally yielded better mean performance than when using the Mahalanobis distance. Figure~\ref{fig:CXRa} shows how the proposed SPC-based method successfully flagged the randomly selected 100 images as OOD; note that each point here represents the OOD metric computed for a single image. 

\vspace{7pt}

To further probe the performance of the proposed method, we conducted further experiments under different OOD scenarios. For the first additional experiment, we conducted subgroup analysis on modality type to report the performance in detecting various modalities individually. Our findings indicate that our method exhibited robust OOD detection across all non-CXR modalities, achieving an accuracy of $0.98$ and a sensitivity of greater than $0.96$. These results suggest that the proposed method is modality-independent. Specifically, our method showcased the ability to flag various imaging modalities (e.g., CT, MRI, bone X-ray, and ultrasound) as OOD compared to the CXR modality. This ability to recognize images as OOD, irrespective of the modality, offers an advantage as it eliminates the need for training separate models for each modality to classify them as OOD. 

In addition to assessing the ability of the proposed method to detect OOD irrespective of the modality, we wanted to assess the performance of the method across different CXR datasets. Therefore, we repeated OOD detection using models trained on the Padchest dataset \cite{Bustos_2020}. PadChest contains over 160,000 images from 67,000 patients, interpreted and reported by radiologists at Hospital San Juan (Spain). Our results showed that $3\sigma$ detection using the Padchest-trained model and the same set of OOD images achieved a sensitivity of $0.982$, specificity of $0.966$, and an overall accuracy of $0.974$. These results support our proposition that SPC-based OOD detection is dataset-independent. SPC techniques, by design, are adaptable to various types of data distributions and can be calibrated to detect deviations from a norm, making them inherently flexible and applicable across different datasets.

\subsubsection{Results for SPC-based OOD Monitoring}
In our CXR monitoring simulation (see Section~\ref{sec:Sim}), we similarly assumed a clinical facility collects and labels 100 radiographs per day. During the first month, we set the daily OOD rate to uniformly vary between 0-1\%. In the second month, we increased the OOD rate to be uniformly sampled from 2-4\%.

Figures~\ref{fig:CXRb} and \ref{fig:CXRc} illustrate the performance of both $3\sigma$ and CUSUM monitoring methods. A comparison of Figure~\ref{fig:CXRa} with Figure~\ref{fig:CXRb} reveals that the $3\sigma$ SPC chart effectively flagged individual OOD images, yet its performance in identifying OOD batches was less efficient, as shown in Figure~\ref{fig:CXRb}. However, it is observable that points post day 30 consistently fell below the mean and approached the control limits. This pattern suggests that these points would be flagged as OOD under different runtime rules, such as if two out of three consecutive points lie beyond the $2\sigma$ limit. In contrast to the daily average $3\sigma$, the CUSUM method, depicted in~\ref{fig:CXRc}, successfully identified the shift that occurred on day 31, with a detection delay of only one day when a k-value of 0.5 was used. See Table \ref{tab:tuningk} in the Appendix for a detailed evaluation of the performance of the CUSUM method across a range of k values.

\begin{figure}
     \centering
         \includegraphics[width=\linewidth]{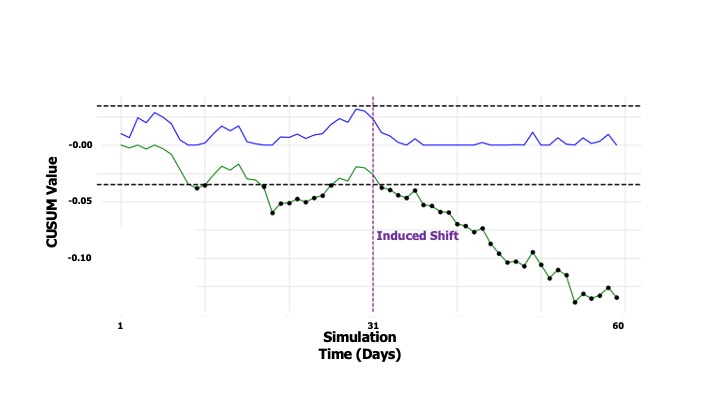}       
        \caption{~\textbf{CUSUM monitoring ($h=0.50$) for distributional shift towards pediatric CXR inputs.} The blue and green signals represent the high- and low-side CUSUM signals, respectively. Black dashed lines indicate $\pm$ thresholds. Black dots are OOD flags. Purple dashed line indicates induced shift.\\ }
     \label{fig:peds_CUSUM}
 \end{figure}

\vspace{7pt}
\textbf{Detecting a Demographic Drift.} We also assessed whether our method could effectively monitor and flag changes in the demographic characteristics of patients' imaging data, such as age group variations. Our results showed that the proposed method can detect pediatric CXRs as OOD. Despite the control limits being set using adult CXR images, our method successfully detected this distribution change, achieving an accuracy of $0.892$, sensitivity of $0.894$, and specificity of $0.891$. We also found that CUSUM rapidly detected the demographic change (Figure~\ref{fig:peds_CUSUM}). However, we did observe several false positive flags prior to the induced shift. This complexity arises partly because the used pediatric dataset includes a wide age range, from infants to 18-year-olds, making some pediatric CXR images very similar to those in the adult training set (see Figure~\ref{fig:PedCXR_Example} in the Appendix). Nonetheless, the proposed method maintains acceptable performance in identifying the induced shift—specifically, the incorporation of 3-5\% pediatric CXR into the daily batch (consisting of 100 images) of adult CXR.

\section{Discussion}\label{Section V}

In this work, we present a framework for monitoring inputs to AI/ML models using Statistical Process Control (SPC). We demonstrated the utility of this framework for detecting individual out-of-distribution (OOD) images as well as monitoring drift in the distribution of input data.

Our investigation showed that successful SPC-based OOD monitoring requires suitable feature representations of the underlying data and OOD metrics. We demonstrated that the choice of features and OOD metrics are task-specific because different clinical scenarios have different sensitivity and specificity operating points as well as different data availability for model training, which is required for defining the in-distribution that sets the control limit. 

We demonstrated these design decisions in two simulated clinical scenarios. In the first, we monitored an input stream of CT images for non-axial slices where in- and out-of-distributions were available to train the monitoring algorithm. In the second scenario, labeled in- and out-of-distrubition examples were not available to train a CXR monitoring algorithm to detect non-CXRs or demographic shifted CXRs in a primarily adult CXR input stream. A central difference between these tasks is the access to supervised labels during training. Feature extraction for the Axial CT monitoring task necessitated training on a repository of slice-supervised CT images; therefore, as we show in Table \ref{tab:CT_results_summary}, the best features for the CT task were derived from slice-supervised contrastive training. In contrast, we did not know \textit{a priori} what the OOD distribution would look like in the CXR task. The OOD images in the test set are comprised of medical images of diverse modalities, including ultrasound, MRI scans, and bone X-rays, as well as pediatric CXR. Thus, the optimal feature space representation relies on training with the disease pathology labels and using a vector similarity metric to flag OOD data (see Table \ref{tab:CXR_results_summary}). Through our series of experiments, we successfully demonstrated that our proposed framework can be applied in both scenarios.

In both tasks, we illustrate that detection and monitoring use different SPC schemes: $3\sigma$ is effective at image-level detection, but CUSUM is better suited for monitoring data drift. With $3\sigma$, we set the mean $\mu$ and standard deviation $\sigma$ of the distance metric in feature space based on the training set. At runtime, images were flagged if their OOD metric exceeded $3\sigma$. This method showed high specificity and sensitivity in individual OOD image detection (Tables \ref{tab:CT_results_summary} and \ref{tab:CXR_results_summary}, Figures \ref{fig:CTa}, and \ref{fig:CXRa}). However, it struggled to flag temporal drift in batched input streams (Figures \ref{fig:CTb} and \ref{fig:CXRb}), as daily averaging of scores tended to obscure anomalous deviations in the distance metrics. In contrast, CUSUM, which accumulates data drift, was more effective in these scenarios (Figures \ref{fig:CTc} and \ref{fig:CXRc}), consistently detecting induced shifts after only a few days. In short, our findings indicate that while $3\sigma$ flagging effectively detects OOD at the image level, CUSUM is more adept at identifying drift in batches of images (e.g., 100 images per day). However, it is important to note that our investigation was limited to the $3\sigma$ rule for SPC. Exploring additional runtime rules (e.g., two out of three consecutive points exceeding the $2\sigma$ limit, 7 consecutive points to one side of the mean) could potentially improve the sensitivity in detecting OOD. This possibility is underscored in Figure~\ref{fig:CXRb}, where following day 30, the data points consistently descended below the mean and neared, but did not cross, the control limits. 

We additionally show how CUSUM monitoring is sensitive to changes in the underlying distribution of patient demographics. Although we did observe some false positive CUSUM flags (Figure~\ref{fig:peds_CUSUM}), we argue this is likely due to the age heterogeneity within the pediatric CXRs. This highlights the importance of precise calibration of CUSUM parameters. For instance, choosing a larger $k$ value might have minimized these false positives, albeit at the risk of increasing the delay in accurately detecting the shift to pediatric CXR images after day 30. This is a critical trade-off in parameter selection: reducing false positives while maintaining timely and accurate detection of true data changes. In future works, we plan to stratify the pediatric CXRs into distinct age groups and evaluate our method's performance within these specific categories. We also plan to explore other demographic characteristics including race and gender. 

Although several methods for OOD detection exist in the literature, the proposed method offers several benefits. First, using SPC charts presents an adaptive and data-driven method for threshold setting. Traditional OOD detection methods require setting fixed thresholds based on assumptions or prior knowledge. Our approach utilizes statistics of the data to determine the threshold, ensuring that the threshold is informed by the data and thus removing human judgment, which can be inconsistent and introduce biases. Further, by plotting metrics over time, we can monitor data changes or drifts over time. Whereas fixed thresholds become obsolete over time, data-driven thresholds from SPC charts can adapt, making our approach more robust to data drift and continuous monitoring.

As SPC methods are rooted in statistical concepts, they are generally applicable to monitoring scenarios where data is approximately Gaussian distributed, which is a reasonable assumption for most real world applications of sufficiently large sample sizes. These simple assumptions enable the application of SPC to a broad spectrum of data sources including CT and CXR monitoring as demonstrated in this work. 

In future studies, we plan to apply our OOD framework to prospective clinical data as well as explore its utility as a data-cleaning tool. The dual capacity of our method to operate on a temporal and per-image (individual image) scale makes both applications possible. We are also interested in evaluating the efficacy of different OOD flagging rules. As we alluded to when comparing $3\sigma$ and CUSUM monitoring, throwing a single flag does not represent a confident OOD detection and might result in frequent false positives. Therefore, additional SPC rules might be more useful in real clinical scenarios.


\section{Conclusions}
\label{sec:Conclusion}

In this work, we proposed a framework for OOD detection and monitoring using SPC methods. We combined machine learning feature extraction methods and geometric-based distances with SPC charts to track deviations from training data distributions. We employ geometric metrics such as cosine similarity and Mahalanobis distance to quantify these deviations and generate OOD metrics. By plotting OOD metrics on SPC charts, we efficiently and promptly flagged deviations from expected distributions. We evaluated our method in two applications: flagging non-axial CT views as OOD, and identifying non-CXR and demographically shifted CXRs as OOD.

Our proposed SPC-based framework has demonstrated promising results across diverse tasks, showcasing its efficacy and potential as a robust approach for AI monitoring in various domains. The proposed framework can be extended to other medical imaging applications as well as other modalities. It can also be used as an approach for open set recognition, where unknown classes are flagged as OOD. Sensitive drift detection can also inform when the base classifier should be retrained. By monitoring for a series of OOD metrics outside the control chart boundaries, we offer a systematic way to decide when retraining (or re-calibration) might be beneficial. That is, when an OOD flag is raised, a clinical team can interrogate the flag and decide whether to initiate retraining of feature extraction models or recalibration of SPC parameters. This would be valuable in clinical settings where human-model feedback is imperative to longitudinal OOD monitoring.

Our results demonstrate the effectiveness of the proposed framework for identifying and correcting OOD errors, such as incorrect CT views and CXR modalities, within large datasets. This capability is not just an academic exercise but a practical tool in addressing the real-world issue of data drift and mislabeling in clinical settings. The mislabeling of exams, as evidenced by previous research can have significant repercussions, including the misapplication of AI models to inappropriate studies. This research, therefore, serves as a bridge between theoretical ML advancements and the pressing needs of the healthcare industry, highlighting the critical role of accurate data labeling and quality assurance in the deployment of AI technologies.

\section*{Acknowledgment}
This project was supported in part by an appointment to the ORISE Research Participation Program at the Center for Devices and Radiological Health (CDRH), U.S. Food and Drug Administration, administered by the Oak Ridge Institute for Science and Education through an interagency agreement between the U.S. Department of Energy and FDA. The mention of commercial products, their sources, or their use in connection with material reported herein is not to be construed as either an actual or implied endorsement of such products by the Department of Health and Human Services. This is a contribution of the U.S. FDA and is not subject to copyright. 

\bibliographystyle{ieeetr}
\bibliography{references}  






\clearpage
\centering
\section*{Appendix}  
\setcounter{page}{1}

\begin{figure}[!h]
\centering
\includegraphics[width=0.50\textwidth]{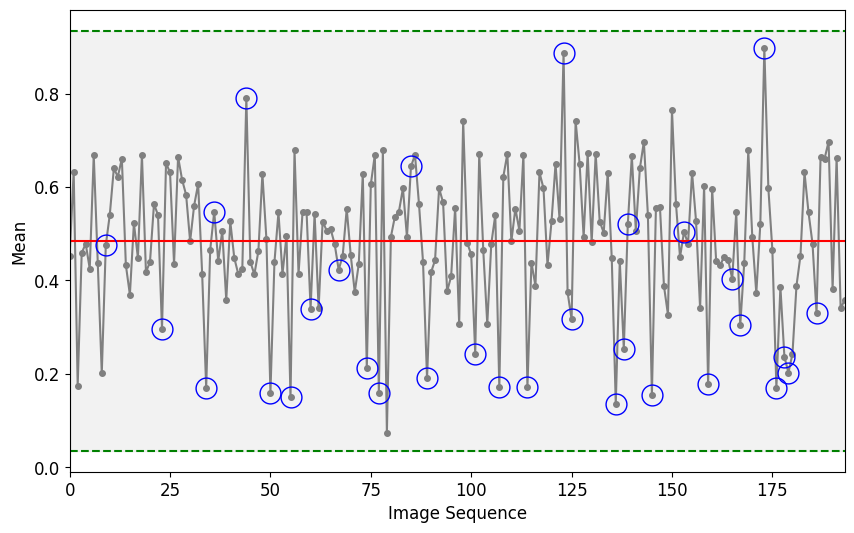}
\includegraphics[width=0.50\textwidth]{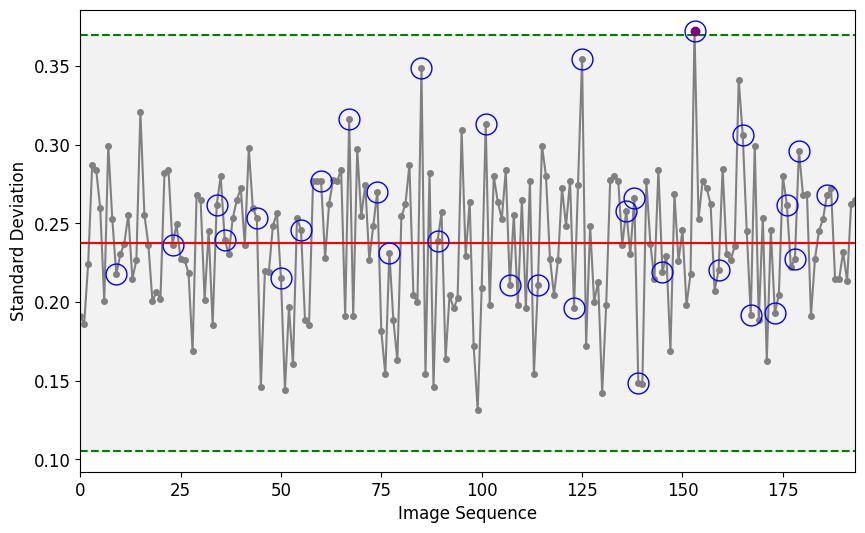}
\includegraphics[width=0.50\textwidth]{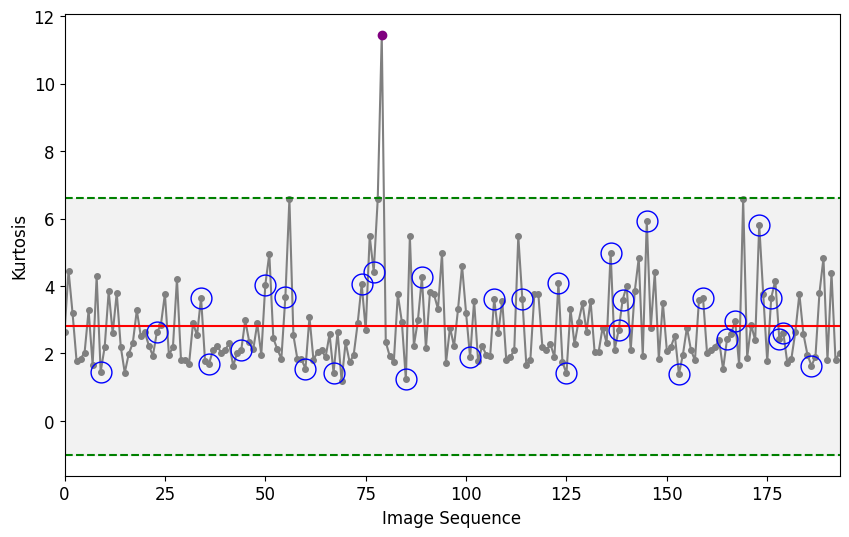}
\includegraphics[width=0.50\textwidth]{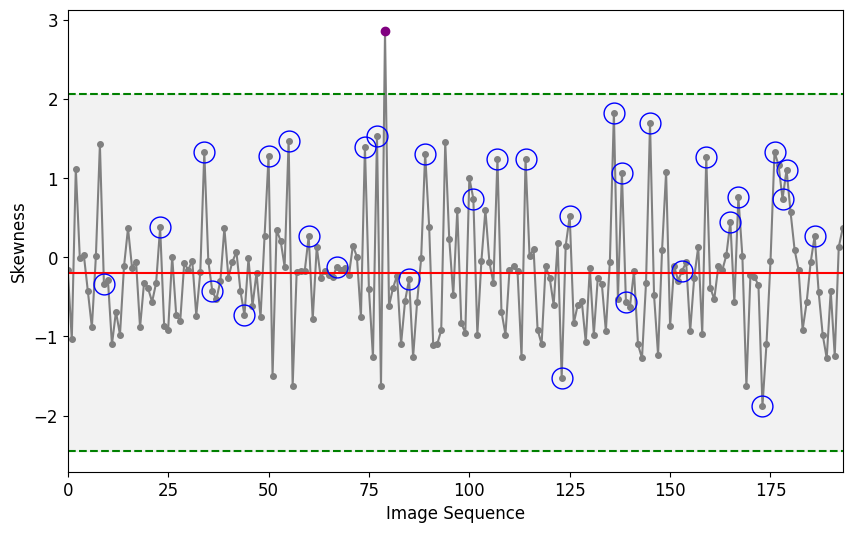}
\caption{~\textbf{$3\sigma$ OOD detection of non-CXRs using simple image statistics.} Four zero-order features (mean, standard deviation, kurtosis, skewness) were evaluated for OOD detection.}

\label{fig:image-statistics}
\end{figure}

\begin{figure}[!h]
\centering
\includegraphics[width=0.70\textwidth]{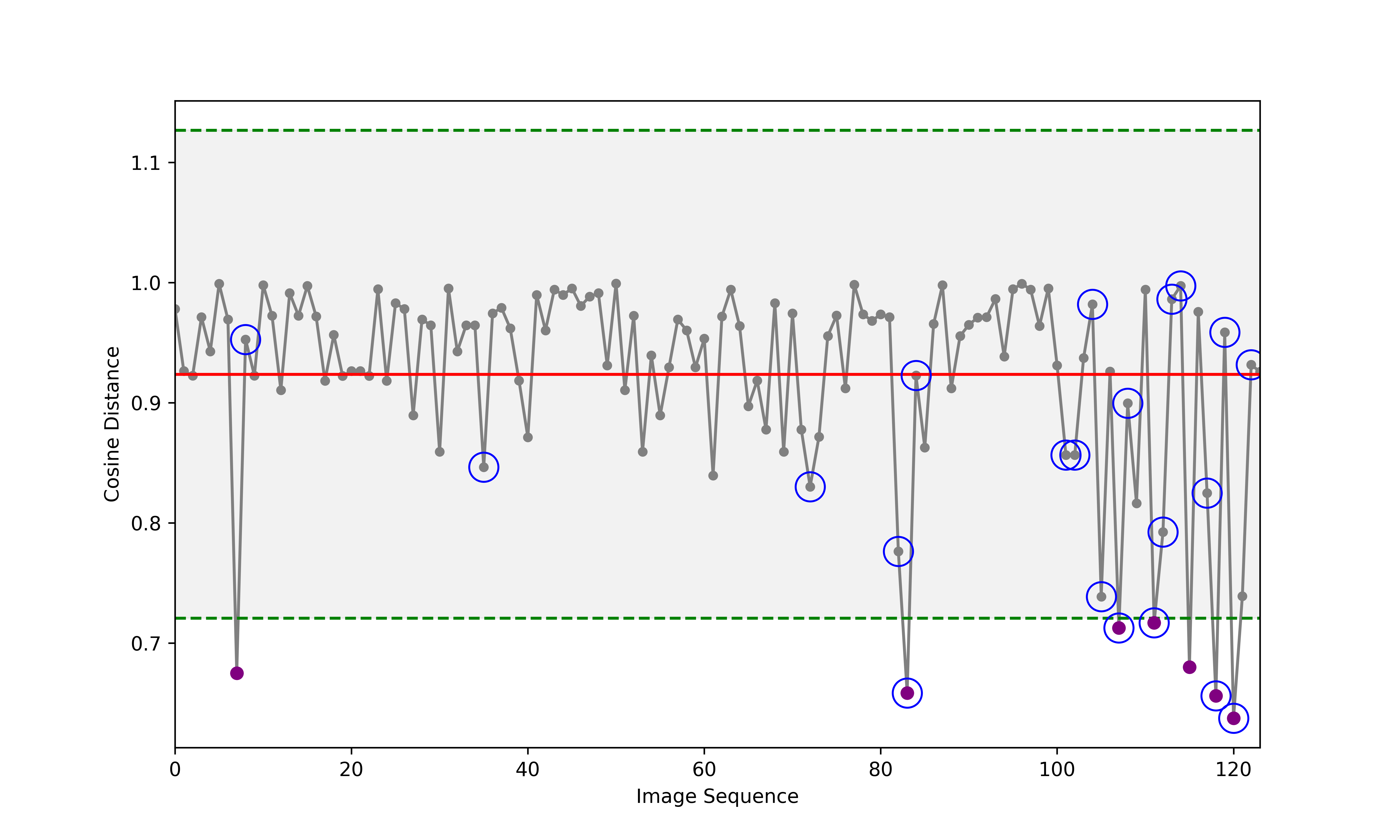}
\caption{~\textbf{$3\sigma$ OOD detection of non-CXRs using texture features.}  We utilized grey level co-occurrence matrices (GLCM) to quantify texture information, recording the frequencies of adjacent pixel pairs in an 8-bit image. By averaging these matrices across the eight nearest neighboring pixels, we accounted for rotational variations. We then derived simple statistics to use as SPC features.}
\label{fig:texture-features}
\end{figure}

\begin{table}[ht]
\centering
\begin{tabular}{|l|l|}
\hline
\textbf{Parameter} & \textbf{Value} \\ \hline
Number of Epochs & 10 (CXR), 100 (CT) \\ \hline
Loss Function & Mean Squared Error Loss \\ \hline
Optimizer & AdamW \\ \hline
Learning Rate & 0.001 \\ \hline
Hidden Channels (c\_hid) & 16 \\ \hline
Latent Dimension & 100 \\ \hline
Encoder Conv Layers & 5 Conv2d, 5 ReLU \\ \hline
Decoder Conv Layers & 5 ConvTranspose2d, 5 ReLU \\ \hline
\end{tabular}
\vspace{6pt}
\caption{~\textbf{Summary of autoencoder parameters for CT and CXR applications.}}
\label{tab:autoencoder}
\end{table}

\begin{table}[!h]
\centering
\begin{tabular}{|l|l|}
\hline
\textbf{Parameter} & \textbf{Value} \\ \hline

Optimizer & Adam (learning rate = 0.0001 (CXR), 0.001 (CT)) \\ \hline
Loss Function & Binary Cross-Entropy \\ \hline
Number of Epochs & 10 (CXR), 100 (CT) \\ \hline
 Batch Size & 32 (CXR), 128 (CT) \\ \hline
\end{tabular}
\vspace{6pt}
\caption{~\textbf{Summary of parameters in supervised binary cross-entropy learning for CT (ResNet-18) and CXR (VGG16) applications.}}
\label{tab:supervised_BCE}
\end{table}

\begin{table}[!h]
\centering
\begin{tabular}{|l|l|}
\hline
\textbf{Parameter} & \textbf{Value} \\ \hline
Loss Function & Supervised Contrastive (SupConLoss) \cite{khosla2020supervised}\\ \hline
Temperature for Loss & 0.07 \\ \hline
Training Approach & Batch-wise training with accumulated loss \\ \hline
Optimizer & Adam (learning rate = 0.0001 (CXR), 0.001 (CT)) \\ \hline
Batch size & 32 (CXR), 128 (CT) \\ \hline
\end{tabular}
\vspace{6pt}
\caption{~~\textbf{Summary of parameters in supervised contrastive learning for CT (ResNet-18) and CXR (VGG16) applications.}}
\label{tab:supervised_contrastive}
\end{table}


\begin{table}[!h]
\centering
\begin{tabular}{|c|c|c|c|c|}
\hline
\textbf{$k$} & Delay (CT) & FP (CT) & Delay (CXR) & FP (CXR) \\
\hline
0.10 & 2 & 0 & 1 & 0\\
\hline
0.25 & 3 & 0 & 2 & 0\\
\hline
0.50 & 4 & 0 & 2 & 0\\
\hline
0.75 & 4 & 0 & 2 & 0\\

\hline
\end{tabular}
\vspace{6pt}
\caption{~\textbf{Impact of CUSUM parameter $k$ on detection delay (measured in days after day 31) and false positives (FP).}}
\label{tab:tuningk}
\end{table}

\begin{figure}
    \centering
    \includegraphics[width=0.90\textwidth]{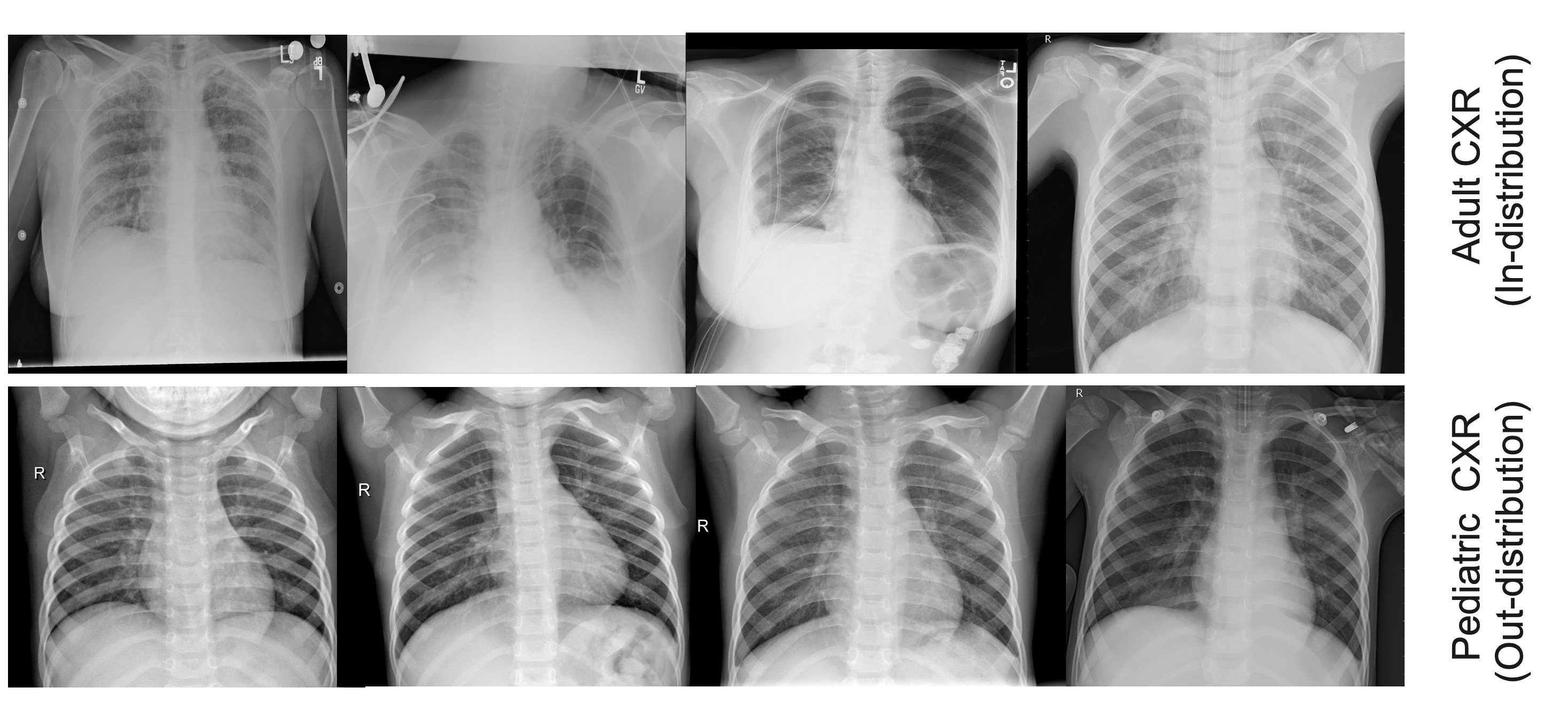}
    \caption{~ Examples of Adult CXR images (in-distribution) and pediatric CXR images (out-distribution).}
    \label{fig:PedCXR_Example}
\end{figure}

\end{document}